# Combining Neuro-Evolution of Augmenting Topologies with Convolutional Neural Networks

Jan Hohenheim, Mathias Fischler, Sara Zarubica, Jeremy Stucki

December 1, 2022

**Abstract**

Current deep convolutional networks are fixed in their topology.

We explore the possibilites of making the convolutional topology a parameter itself by combining NeuroEvolution of Augmenting Topologies (NEAT) with Convolutional Neural Networks (CNNs) and propose such a system using blocks of Residual Networks (ResNets).
We then explain how our suggested system can only be built once additional optimizations have been made, as genetic algorithms are way more demanding than training per backpropagation.

On the way there we explain most of those buzzwords and offer a gentle and brief introduction to the most important modern areas of machine learning.

# Contents

# 1 Introduction to neural networks

## 1.1 What is a neural network?

The most famous neural network is you. Or, in other words, the human brain.

It is, simply put, a clever arrangement of smallest units capable of processing easy logic.

These smallest units are called **neurons**, and our brain consists of approximately 100 billion of them. They are interweaved through a complex series of incomming and outgoing extensions called dendrites and axons, respectively, of which some transport electricity faster than others. Most of the components of a brain are unfortunately still not understood well enough to be used productively in computer science.

An **artificial neural network** (ANN) tries to emulate the immense success of its biological counterpart by abstracting the complex chemical reactions responsible for our thoughts to much more graspable math. The feedforward version of such an ANN consists of two simple components: neurons and connections.

Each neuron has inputs, which are the incoming connections. It applies a simple mathematical operation to this set of inputs and returns the result.

Connections connect neurons to each other. Each connection has a weight, which determines how weak or strong the connection is.

The neurons are typically organized into layers. The first is referred to as the input layer and the last one as the output layer. The remaining layers are called hidden layers. (Anderson(1995))

Here is a basic example of a neural network:

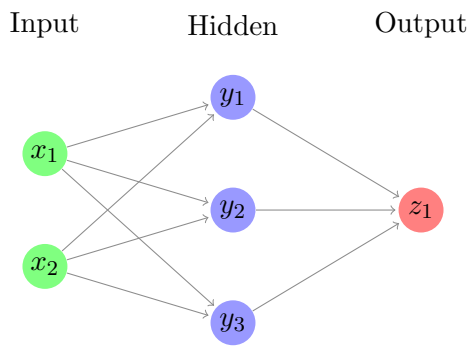

Each connection is represented as an arrow and has an associated weight. Every neuron is connected to all neurons in the previous and in the next layer.

The configuration of how all neurons and layers are interconnected, as well as the number of layers, is called the **topology** of the network.(Anderson(1995))

For simple networks, you can also write down the inputs and the corresponding outputs.

$$\begin{vmatrix} 0 & 0 \\ 0 & 1 \\ 1 & 0 \\ 1 & 1 \end{vmatrix} \rightarrow \begin{vmatrix} 0.03 \\ 0.76 \\ 0.87 \\ 0.10 \end{vmatrix}$$

This network was trained to solve the XOR problem, which can be simplified as "are my inputs different?". We defined the output to represent *yes* if its $>= 0.5$ and *no* otherwise.

It's also possible for a neural network to have multiple outputs.

We will use the assumption that our network has one output per possible answer for the rest of the documentation.

Example: We have a picture of a flower. It can either be a poppy, a lilly or a dandelion. Our neural network looking at the flower would have three outputs.

In this case we still wish to have one definitive output. For this we'll use the **softmax** function, which squashes all our outputs in matter that lets them add up to exactly one. One can think of it as a normalization that represents confidence.(Anderson(1995))

It is defined as follows:

$$\sigma(\mathbf{z})_j = \frac{e^{z_j}}{\sum_{k=1}^{K} e^{z_k}}$$

for j = 1, ..., K.

Example: If we our outputs are $[1, 2, 3, 4, 1, 2, 3]$, the softmax of that is $[0.024, 0.064, 0.175, 0.475, 0.024, 0.064, 0.175]$

We then simply take the highest one as our main output. This is called the *winner takes all principle* and is modeled after how the brain works (Anderson(1995))

## 1.2   How does a neural network learn

### 1.2.1   Traditional

A traditional approach of optimizing the connection weights to improve the network's accuracy is named backpropagation.

> *"The Backpropagation algorithm is a supervised learning method for multilayer feed-forward networks from the field of Artificial Neural Networks.*
>
> *Feed-forward neural networks are inspired by the information processing of one or more neural cells, called a neuron. A neuron accepts input signals via its dendrites, which pass the electrical signal down to the cell body. The axon carries the signal out to synapses, which are the connections of a cell's axon to other cell's dendrites."* (Brownlee(2016))

The backpropagation algorithm is a algorithm for supervised learning. In supervised learning, it is being measured how good a network performs, by testing a network with a given dataset, over and over again.

In such a dataset, input values and the expected outputs for these values are defined.

The discrepancies from the specified outputs in the dataset and the actual outputs are called the *errors* of the network.(Polamuri(2014a))

Using basic calculus, the so called *error* of a network can be calculated. This is also known as solving the error minimization problem. (Sathyanarayana(2014))

> *"In the most popular version of backpropagation, called stochastic backpropagation, the weights are initially set to small random values."*(Sathyanarayana(2014))

Stochastic methods are being used, because *"properly scaled random initialization can deal with the vanishing gradient problem"*(Philipp Krähenbühl(2016))

With enough complexity, neural networks can represent any existing function. (Nielsen(2016))

There are methods for picking initial weights, so that problems with local maximums of derivatives are not limiting the backpropagation algorithm.(Nguyen & Widrow(1990))

However, as Dr. Geoffrey E. Hinton states, backpropagation is often limited by the sheer sizes of networks that are required today:

>"Backpropagation was the first computation- ally efficient model of how neural networks could learn multiple layers of representation, but it required labeled training data and it did not work well in deep networks." (Hinton(2007))

### 1.2.2   Genetic algorithm

The training starts with a number of genomes, typically referred to as the population. For each of these genomes a network is built and it is tested against the expected outputs. From these results we can assign a fitness to the genome. A higher fitness indicates that the genome was able to solve a problem better than another. (Anderson(1995))

The initial set of genes is the first generation. The weights of all genes are set to a random value.

To get to the next generation, all genomes have to be tested. Before that, each genome has a chance that a random gene mutates, E.g. the gene is assigned a new random weight.

After that, we select the genomes for the next generation. To select a genome, a so called roulette wheel selection is performed. This means that every genome has a chance to get to the next generation, based on its fitness. (Bäck(1996))

We always select two genomes at a time, so that we can perform a crossover. This means that we swap a part of the genes in the first genome with the second. (Buckland())

This process is repeated until a genome reaches the target fitness, which is set by the trainer.

## 2 What is NEAT

NEAT stands for Evolving Neural Networks through Augmenting Topologies and is a technology first proposed by O. Stanley. (Stanley(2002))

It presents an elegant way to combine genetic algorithms with evolving topologies.

### 2.1 Topology

In traditional neural networks, the topology is fixed. The number of hidden layers and the number of neurons in each hidden layer are given. This makes it very easy to see the difference between two networks, since the only differences are the weights.

The downside is, that the performance of these networks heavily depends on the chosen topology, which leads to the conclusion that many networks would perform better if one had chosen a different topology.

NEAT proposes a technique to evolve the topology over time which allows the network to be better structured for a specific task then a configuration with hyper parameters.

The main problem of such a network, called *Topology and Weight Evolving Artificial Neural Network*, or TWEANN for short, is the *competing conventions problem* (Stanley(2002)). It means that two networks may generate the same solution to a problem at different points in time, thus appearing to be two distinct topologies.

This makes the algorithm mark them as not compatible for a genetic crossover during the mating phase.

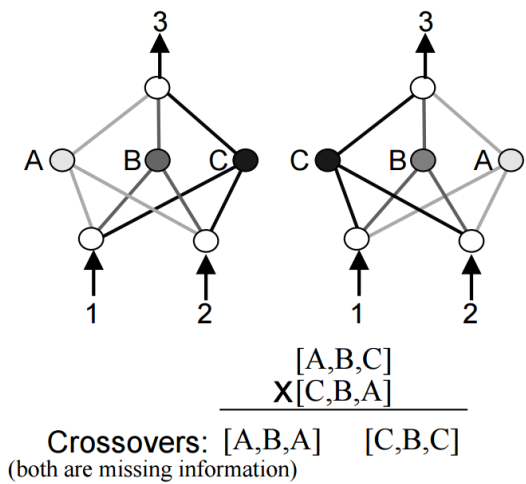

NEAT solves this problem by assigning each connection a *historical marking*, which can be imagined as a serial number.

The first gene ever created is corresponds to a historical marking of one, the next one to two, and so on. Every new gene is then first compared with all existing genes. If an identical match is found, the new gene gets the same historical marking as its twin. If not, the next biggest total number is assigned to it.

This way, during crossover, the algorithm doesn't have to check any complicated structural compatibility, but instead simply compares the historical markings of the two networks. If they are largely the same, the networks are suitable for a genetical exchange.

## 2.2 Speciation

Another difficulty in evolving topologies lies in the way the topology is encoded in the genome. When a new connection is introduced in a network, it's often first a bit worse than before because it needs some time to adjust and show it's real potential. Traditional TWEANNs like to throw these kinds of topologies out of the gene pool preemptively, as they *appear* to make the network worse.

NEAT solves this by again by using historical markings. The more markings a network shares with another, the more related it is to that other network. Based on this principle, NEAT groups similar networks into species, which share their fitness with each other. This means that weak individuals that are only marginally different from a proven concept are guaranteed to be temporarily protected in their niche. (Stanley(2002))

# 3 Convolutional Neural Networks

## 3.1 Problems with image recognition

Most neural networks are unable to handle the amount of data contained in an image. For example an image with a resolution of 3264x2448 (8 Megapixels) would result in almost 24 million inputs, as each pixel is split into its red, green and blue parts.

Another challenge is the detection of so called "features" across an image. Traditional neural networks only detect a feature at a specific location in the image. This is a big issue in image recognition, as you almost always want the entire image to be handled equally. A self-driving car should recognize a stop sign, regardless of its position in the image.

## 3.2 Subsampling

Subsampling is, broadly speaking, the act of taking values from a source, observing them and combining these into a smaller dataset that is still representative.

It's a bit like compressing, really.

Traditional use cases of subsampling include the JPEG format. It makes use of the fact that the human eye cannot differentiate colors as good as luminance, and simplifies parts of the image that are not differentiable for the average human anyways. (van den Branden Lambrecht(2001))

### 3.2.1 Kernels

The subsampling CNNs perform is not related specifically to the human eye but animal visual systems in general. (Masakazu Matsugu(2003))
The main goal of a CNN is to "see" structures in images. These can be geometric (line, square, circle, etc.), typical human recognitions (face, smile, house, cat), and also totally inhuman and unintuitive structures (wiggly lines pointing to the left, three stripes ending up in a sharp point)

Enter kernels.
Kernels are little matrices (rectangular tables of numbers) that go through an image and filter a certain structure out of it as they multiply their weights with the individual pixels. Because of this behavior, they are sometimes referred to as **filters**.

An aggregation of filters with an equal size is called a **convolution**.

When working with convolutions, we refer the inputs and outputs as **tensors**.

A tensor is, in layman's terms, a matrix with more than two dimensions. A tensor with three dimensions, which is called a *tensor of rank three* in maths, can be imagined as a cube.

A convolution takes a tensor of variable dimensionality as an input and returns a tensor of rank $n$, where $n$ equals the number of filters in the convolution. The exact size of the input tensor is irrelevant, as the convolution reapplies its filters over the whole input.

### 3.2.2 Poolers

Despite the kernels doing a great job at making the image smaller, the resulting data is still quite too big to work with. For that reason one can use poolers, which are nothing but dull compression algorithms. The most used pooler is the max pooler. (Graham(2014))

This simple unit traditionally takes four adjacent pixels, then determines the darkest one, and simply concatenates the four original pixels into this smaller single one. Repeat this process over the whole image, and you just scaled it to one fourth of it's original size.

### 3.2.3 Activation function

Every procedure and concept that we described so far ia a linear function. To make a combination of layers meaningful, we need to introduce nonlinearities after each convolution, as a stack of layers would otherwise behave like just one big linear layer.

This is done by an activation function layer.

The most commonly used one in the field of image recognition is the rectifying linear unit, in short **ReLU** (Alex Krizhevsky & Hinton(2012)). It's definition is extremely easy:

$$f(x) = max(x, 0)$$

In other words, it just replaces every negative value in a feature map for a zero.

# 4 Hippocrates, a NEAT implementation

## 4.1 Motivation

The currently available implementations of NEAT are suboptimal.

Most machine learning frameworks and libraries are focused on training by backpropagation and only offer limited support for genetic algorithms.

Dr. Stanley's original implementation in C++ (Stanley(2010)) was written before the major revisions in the C++ language, which made the language very different to use. (Stroustrup(2013))

The Code is no longer effectively usable, as it is ridden with experimental features, afterthoughts, dead code and patterns of thought that are no longer in use.

The most usable implementations are all written in python, which makes them very easy to use but also very slow when compared to optimized C++.

This is why in 2014 Mr. Ferner decided to work on an "actually usable" implementation of NEAT, which he called Hippocrates.

## 4.2 Technology

At the beggining, a big question was, in which language we should write our library in.

The main contestants where C++ and C#. We had to juggle different pros and cons.

One aspect is, how easy the actual writing would be. There is a concept in programming languages which is called *memory safety*. It describes, how and when objects end their accessibility.

Just like real life, a program is made out of various objects, each of which having a distinct state and possible actions.

One such possible object could be a dog. It's state, which is divided into a set of *variables*, could for example consist of his age, his haircolor, his character and so forth.

His possible actions, which are called *functions* in the programming world, could include bark, walk or lick. Some of his functions might even alter his state, like a function for celebrating birthday might change the age variable by +1. These objects that compose a programm however have to, just like in real life again, die off at some point.

Our program might spawn hundreds or thousands of objects. If we do not do something about it, these objects would clog up our entire memory and slow every process down. The question becomes "when does their lifetime end"?

So called safe languages like C# answer by saying "whenever absolutely no one needs them anymore anywhere". This very hedonistic principle is enforced by a *garbage collector*. This is a program that carefully inspects a running process and it's objects and finds out, if an object is really not used anymore. Modern day garbage collectors have become very efficient at what they're doing, but still require performance. Another disadvantage is also that garbage collectors are non deterministic, which means that a programmer can never now for sure at what exact point the objects get destroyed. If the garbage collector that it's time to free up some space, it's gonna do it no matter what. If this happens during a performance critical part of the application, it's going to be slowed down by a lot.

The counterpart are unsafe languages. C++ is called unsafe because before 2011 it didn't have a standard way to manage lifetimes of objects except for forcing the programmer to watch over the memory manually, often leading to corrupt data and undefined behaviour during the runtime. (Stroustrup(2013))

In modern C++ however, lifetimes of complex objects can be managed by so called *smart pointers*, which are implemented as reference counters.

This means that everytime a function tries to use an object, it's reference counter goes up by one. If the function is done with it and doesn't need the memory anymore, this counter goes down by one. As soon as the reference counter hits zero, the object is destroyed.

This gives the programmer determinism, which means that he now knows exactly when the memory is going to get freed (provided he designed his application carefully). This however comes at the cost of requiring more design skill than a using a garbage collector.

In certain edge cases it is possible that reference counters use up more performance than a garbage collector, as the latter is free to do more optimizations on the final code provided he can prove that the end effect is the same.

Additional considerations are that the most used machine learning libraries are written in C++, however C# has way better system of actually distributing the libraries.

This gives us a though decision: Do we want the comfort and stability of C# for increased productivity or the absolute control and performance power of C++?

In the end, Hippocrates was written in C++, as we deemed the performance of the library to be of crucial importance to the usability in the future.

Mr. Ferner and Ms. Zarubica already wrote C++ since years at his company, the Messerli Informatik AG, and Mr. Ferner had a lot of experience teaching apprentices the ins and outs of the language, which is why he was happy to assist Mr. Fischler and Mr. Stucki in learning the common syntax and semantics of modern C++.

## 4.3   Discrepancies

### 4.3.1   Paper

To determine if two organisms are compatible for reproductions with each other, one measures the difference of their genomes by a distance function. The original paper describes it as follows:

> *Therefore, we can measure the compatibility distance $\delta$ of different structures in NEAT as a simple linear combination of the number of excess $E$ and disjoint $D$ genes, as well as the average weight differences of matching genes $\overline{W}$ , including disabled genes:*
>
> $$\delta = \frac{c_1 E}{N} + \frac{c_2 D}{N} + c_3 \cdot \overline{W}$$
>
> *The coefficients $c_1$, $c_2$, and $c_3$ allow us to adjust the importance of the three factors, and the factor $N$, the number of genes in the larger genome, normalizes for genome size*

Typical settings for the coefficients are $c_1 = 1.0, c_2 = 1.0, c_3 = 0.4$.

(Stanley(2002))

However, if we look at Stanley's code (Stanley(2010)), the actual formula we find is

$$\delta = c_1 E + c_2 D + c_3 \cdot W$$

Where $W$ is the sum of absolute weight differences.

The same function is used by all the NEAT implementations that we looked at. This deviation is most likely intentional, although not explicitly documented by Stanley himself. In the original function, the importance of excess and disjoint is limited to the sum of 1 (because there can be at most $N$ not matching genes). This means that for two completely different networks, our function results in

$$\delta = 1 \cdot W$$

where $W$ is unlimited. This means, that the weight differences would be a lot more important than the topological ones, which stands in contrast to the usage of the function as an indicator of topological compatibility. (Green(2009))

Because of this, we use the second version of the function without normalization.

### 4.3.2  Original implementation

We didn't implement the ability for neurons to form recurrent connections, i.e. connect to previous layers. This feature is traditionally used to simulate short-term memory in e.g. speech recognition, where one word alters the meaning of another. (Haşim Sak(2014))

As our images are not sequentially interconnected (as e.g. in a movie), we do not need this.

## 4.4  Visualizing Neural Networks

### 4.4.1  Traditional Neural Networks

In the following section, visualizing is meant to be about visualizing the structure, and not about visualizing what neural networks see.(Jason Yosinski & Lipson(2015))

Traditional neural networks are relatively easy to visualize.

An example of this is such a network with two inputs, three hidden neurons and one output neuron.

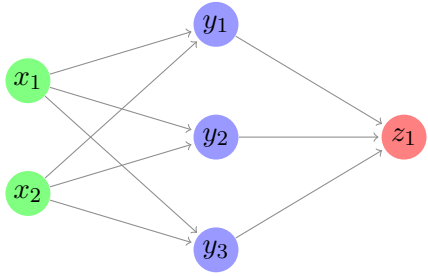

For us, the minimum of visible structure for a neural network to be readable is knowledge of the input layer location, the neurons (displayed as circles) and the connections, displayed as lines.

Additional information that we found useful in understanding the network would be showing the exact weight of the connections.

We created a algorithm to calculate the scaling of the differently sized networks automatically for a fixed space:

For this algorithm, we need information about the amount of layers, the max amount of neurons in any layer.

For the definition of the algorithm, we simply assume the the network will be shown from left to right, with all input-neurons to the left.

The x-axis below is defined horizontal, the y axis vertical.

$width$ is the available width (also $xSize$), $height$ is the available height (also $ySize$).

$layerCount$ is defined as the amount of layers, $maxNeuronCount$ as the max amount of neurons in any layer.

A $Step$ (x or y) defines the distance between the centers of neurons, in x or in y direction, respectively.

$$xStep = (xSize - ((layerCount + 2) * (minMargin + neuronRadius * 2)))$$

$$yStep = (ySize - ((maxNeuronCount + 2) * (minMargin + neuronRadius * 2)))$$

whereat $minMargin$ is the margin that should be kept between neurons to make sure they aren't overlapping, and $neuronRadius$ is the radius of the neurons to be drawn.

$layerCount + 2$ is there, because there are also borders to be kept at the corner of the drawing area to be drawn on - exactly 2 per dimension.

Of course, this means that if $(minMargin + neuronRadius) * (layerCount + 2) > width$ the neurons will overlap anyways and the structure will be hard to read.

This will result in such a structure (taken straight from our software NEAT_Visualizer):

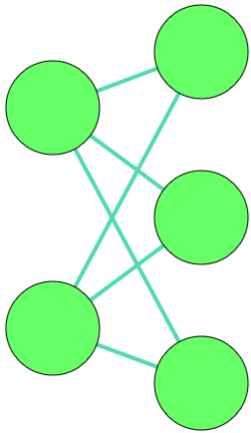

### 4.4.2 Navigating through generations and species

For us, not only the end result was interesting when analysing results from a run, but also the evolvement itself. However, inspecting the evolvement is very interesting, but also complex.

For every generation, there are multiple species, who in turn contain multiple organisms themselves.(Stanley(2002))

In our application for inspecting these structures, NEAT_Visualizer, we can read a JSON dump with logging data from Hippocrates and they will be loaded into the application. The user sees the interfaces as follows:

Left view:

| Gen Index | Fitness | | Species Index | Fitness | | Organism Index | Fitness |
|---|---|---|---|---|---|---|---|
| 1 | 3 | | 0 | 2 | | 0 | 3 |
| 2 | 3 | | 1 | 3 | | 1 | 2 |
| 3 | 3 | | 2 | 3 | | 2 | 2 |
| 4 | 3 | | 3 | 2 | | 3 | 2 |
| 5 | 3 | | 4 | 3 | | 4 | 1 |
| 6 | 3 | | 5 | 3 | | 5 | 2 |
| 7 | 3 | | 6 | 3 | | 6 | 2 |
| 8 | 3 | | 7 | 3 | | 7 | 2 |
| 9 | 2 | | 8 | 3 | | 8 | 2 |
| 10 | 3 | | 9 | 3 | | | |
| 11 | 3 | | 10 | 3 | | | |
| 12 | 3 | | 11 | 3 | | | |
| 13 | 3 | | 12 | 3 | | | |
| 14 | 3 | | 13 | 3 | | | |
| 15 | 3 | | 14 | 2 | | | |
| 16 | 3 | | 15 | 2 | | | |
| 17 | 3 | | 16 | 2 | | | |
| 18 | 3 | | 17 | 2 | | | |
| 19 | 3 | | 18 | 3 | | | |
| 20 | 3 | | | | | | |
| 21 | 3 | | | | | | |
| 22 | 3 | | | | | | |
| 23 | 2 | | | | | | |
| 24 | 2 | | | | | | |
| 25 | 3 | | | | | | |
| 26 | 3 | | | | | | |
| 27 | 3 | | | | | | |
| 28 | 3 | | | | | | |
| 29 | 3 | | | | | | |
| 30 | 2 | | | | | | |
| 31 | 2 | | | | | | |
| 32 | 2 | | | | | | |
| 33 | 3 | | | | | | |
| 34 | 2 | | | | | | |
| 35 | 2 | | | | | | |
| 36 | 3 | | | | | | |
| 37 | 3 | | | | | | |
| 38 | 2 | | | | | | |
| 39 | 3 | | | | | | |
| 40 | 3 | | | | | | |
| 41 | 2 | | | | | | |
| 42 | 3 | | | | | | |
| 43 | 3 | | | | | | |

Right view:

| Gen Index | Fitness | | Species Index | Fitness | | Organism Index | Fitness |
|---|---|---|---|---|---|---|---|
| 1 | 3 | | 0 | 3 | | 0 | 3 |
| 2 | 3 | | | | | 1 | 2 |
| 3 | 3 | | | | | 2 | 2 |
| 4 | 3 | | | | | 3 | 2 |
| 5 | 3 | | | | | 4 | 3 |
| 6 | 3 | | | | | 5 | 1 |
| 7 | 3 | | | | | 6 | 3 |
| 8 | 3 | | | | | 7 | 2 |
| 9 | 2 | | | | | 8 | 2 |
| 10 | 3 | | | | | 9 | 2 |
| 11 | 3 | | | | | 10 | 2 |
| 12 | 3 | | | | | 11 | 3 |
| 13 | 3 | | | | | 12 | 3 |
| 14 | 3 | | | | | 13 | 3 |
| 15 | 3 | | | | | 14 | 2 |
| 16 | 3 | | | | | 15 | 2 |
| 17 | 3 | | | | | 16 | 2 |
| 18 | 3 | | | | | 17 | 3 |
| 19 | 3 | | | | | 18 | 3 |
| 20 | 3 | | | | | 19 | 2 |
| 21 | 3 | | | | | 20 | 3 |
| 22 | 3 | | | | | 21 | 2 |
| 23 | 2 | | | | | 22 | 3 |
| 24 | 2 | | | | | 23 | 1 |
| 25 | 3 | | | | | 24 | 3 |
| 26 | 3 | | | | | 25 | 3 |
| 27 | 3 | | | | | 26 | 2 |
| 28 | 3 | | | | | 27 | 1 |
| 29 | 3 | | | | | 28 | 2 |
| 30 | 2 | | | | | 29 | 3 |
| 31 | 2 | | | | | 30 | 2 |
| 32 | 2 | | | | | 31 | 2 |
| 33 | 3 | | | | | 32 | 3 |
| 34 | 2 | | | | | 33 | 2 |
| 35 | 2 | | | | | 34 | 2 |
| 36 | 3 | | | | | 35 | 2 |
| 37 | 3 | | | | | 36 | 2 |
| 38 | 2 | | | | | 37 | 2 |
| 39 | 3 | | | | | 38 | 1 |
| 40 | 3 | | | | | 39 | 2 |
| 41 | 2 | | | | | 40 | 2 |
| 42 | 3 | | | | | 41 | 3 |
| 43 | 3 | | | | | 42 | 3 |

Both - the left and the right - views are showing a selected generation, a selected species and no organism selected yet. The control reads from left to right - generation, species, organism.

The numbers are always - left first - Index, then Fitness. As an example, the left picture above has generation 15 with a maximum of 3 fitness selected, and its species number 12 with a maximum fitness of 3.

Here is a full view of the visualizer's graphical user interface (and the network drawing algorithm configured to have all the inputs at the bottom):

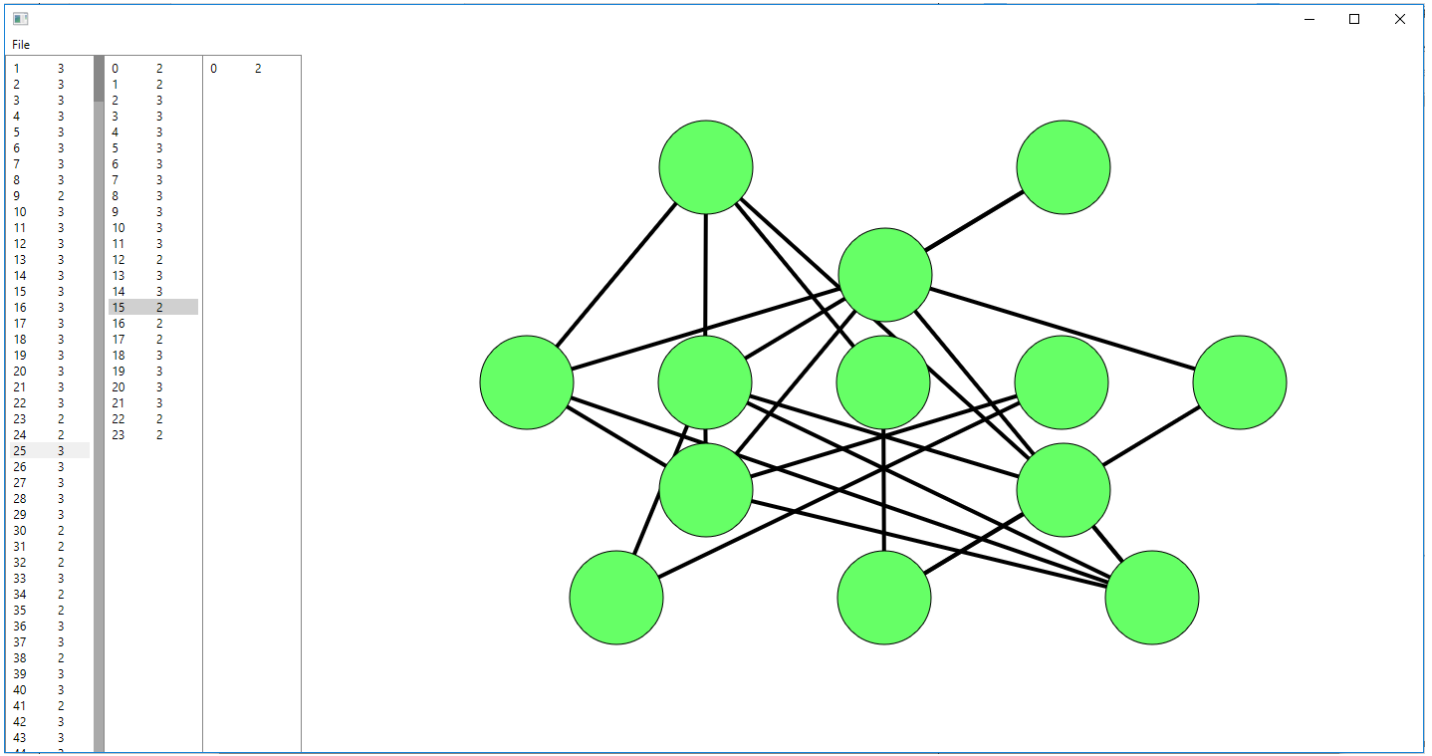

### 4.4.3 Convolutional Neural Networks

*"To visualize the function of a specific unit in a neural network, we synthesize inputs that cause that unit to have high activation."* (Jason Yosinski & Lipson(2015))

This way, a artificial picture can be created that represents what the networks "sees".(Jason Yosinski & Lipson(2015)), (Karen Simonyan(2014))

This is an example of such artificially created images:

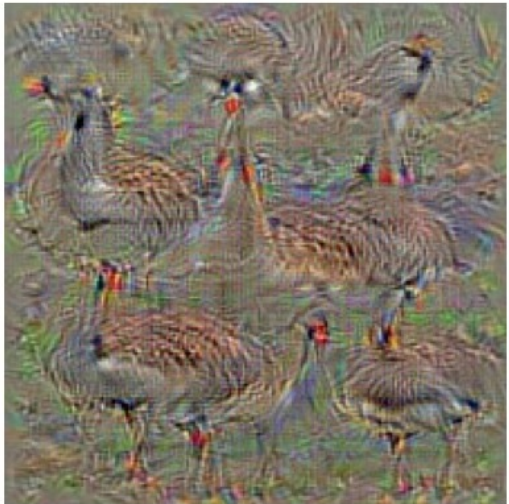 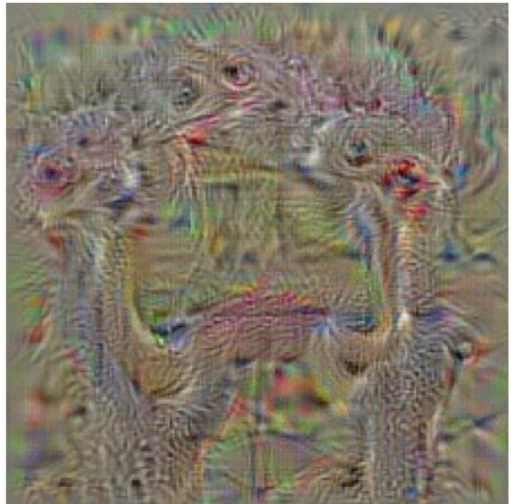

**goose**           **ostrich**

### 4.4.4 Technolgies used for our visualizer

To create a visualizer, we had to chose technologies - for code and graphics.

Due to convenience we decided quite immediately, that C# would be our language of choice. It offers a high productivity with a very concise and remarkable syntax, and is very well known for some of our team users. C# runs on the most used operating systems easily

Also, C# offers a very healthy ecosystem that allows developers and engineers to chose freely between competing products, all more often than not for free.

The decision about what graphics/GUI system to be used was harder.

The prime choice would have been WPF, however, it is limited through it depending on Windows drivers for DirectX. This rules WPF out, because we are convinced of the idea, that if possible, our tools should be available for everyone, not only just Windows users.

Other possibilities would include Gtk-Sharp, WinForms and Avalonia.

The latter one is just in Alpha and was discovered by Mr. Fischler while researching possibilities.

However, it seemed to have similar features and approaches as WPF.

Gtk-Sharp has many appealing features, but no good scalable drawing area. It runs well on Windows, Linux and macOS.

WinForms is very stable due to its age, but will only run on Linux with help of a simulator called Wine. Wine can be found under https://www.winehq.org/.

With that, it seemed the most exciting and still best option to chose Avalonia for development.

Avalonia has a interesting modular system of rendering subsystem, currently supporting Gtk and Cairo (Windows, Linux, macOS) and Win32 with Direct2D (only windows). Skia is currently planned to be implemented to be and replacing Gtk.

### 4.4.5 Interoperability

While creating a tool to visualize the structures of NEATly generated network, we faced multiple problems.

We decided to use Avalonia (https://github.com/AvaloniaUI/Avalonia) as a framework for the visualizer.

Because Avalonia is based on C# (that is a non-native, safe, just in time compiled language), it can not natively exchange data with Hippocrates, that is written in C++ and compiled for a certain platform.

There is a method called interop marshalling, that would provide a solution to such a problem.(Microsoft(2016))

This method however has been designed for Windows, and will work differently on Linux.

Also, it would make the implementation of the visualizer dependent of the memory layout, which is a huge constraint to be taken into consideration. That's why we decided against using interop marshalling.

Another possibility is using the file system to exchange data. All data that belongs together will be contained in a folder with a file per logical unit that it represents.

This is the approach we took to avoid having memory incompatibilities For saving data in a file however, you need a common representation for the data you want to exchange between programs.

The keyword to that problem is serialization. Serializing data is the procedure of converting data from the native memory to a more general (human readable or non-readable) format.

We decided to use a humanly readable and well known serialization format for Hippocrates, because it allows us more flexibility and automation in terms of serializing and deserializing (reading the data into native memory again, but maybe differently structured).

The two most often used and most famous humanly readable data formats are XML and JSON.(Strassner(2015))

We decided to go with JSON, because it is more lightweight and by now more often used than XML.

Memory overflow is a problem when reading lots of data from a file system - it can be fought by not reading all the data, but some data after another and only when needed, and discard as much as possible when not required anymore.

This is often also called *lazy programming* or *lazy initialization*, and it ended up being what we implemented to ensure that the visualizer wouldn't collapse under big Hippocrates dumps.

# 5 Build tools

## 5.1 Version control

A version control system is a computer program that tracks every file change in a directory. It allows to revert to another version of a file if one want to undo something. It is also great for collaboration, as it records who made which change.

### 5.1.1 Git

Git is a version control system that was first proposed by Linus Torvalds in 2005. (Torvalds(a))
It is a free, open-source version control system, which we use for our entire source code and documentation.

We first separated our code into multiple repositories (Torvalds(c)) as we thought it would make sense to keep NEAT and CNN separated. Later we decided that it would make more sense to keep everything in a single repository, as we had to use both parts simultaneously.

A repository is like a project folder, but it is synced across multiple computers.

Git has many powerful tools that defeat their antecessors from other version control by a big margin in terms of usability, performance and stability.

One of these tools is the merge tool. It allows to either automatically - if no conflicts happen - or manually - merge together files from different branches or repositories. This is very useful when working together with multiple teammembers, because you don't have to watch out too much about working in the same files - as long as there's no redundant work done - because the merge tool is able to often fix alot of collisions automatically, or if not, it marks the colliding parts so users have less hard times fixing the conflicts.

When creating and pushing commits onto git repository (a commit is a subset of changes) everyone gets a copy of this commit, as soon as queried for it via "pulling" (getting the latest changes from a remote repository.)(Torvalds(a))

Because of that commit messages are important. They ought to explain what the commit changed on the repository.

To make sure everyone can understand what has been done, we adopted some rule set for naming commits(Beams(2014)):

- Separate subject from body with a blank line

- Limit the subject line to 50 characters

- Capitalize the subject line

- Do not end the subject line with a period

- Use the imperative mood in the subject line

- Wrap the body at 72 characters

- Use the body to explain what and why vs. how

The *limit the subject line to 50 characters* rule is very useful:

It guarantees that on github, the commit message will be readable without requiring a user to expand a area of the page.

The *use the imperative mood in the subject line* rule is useful because it makes commits more readable. As we have used this rule, it has become more and more clear to us that not using imperative means having redundant characters.

As an example, instead of "Add Implementation" the commit message could be "Added Implementation". That is two characters more without any gain of insight or readability. Thats why we found this rule particularly useful.

### 5.1.2  GitHub

GitHub is a web-based Git repository hosting service. (Preston-Werner(a))
This means for us, that we have a central place where our data is located. This also enables us to simultaneously work on the code, which increases the speed of development.

GitHub provides all services for free when developing an open-source application. (Preston-Werner(b))

We use some of the GitHub features to improve the quality of our documentation. We have set it up, so that any change to the documentation has to happen on a new branch and before it gets copied over to the main one, another member of the team has to approve the changes. (GitHub, Inc.(a))

## 5.2 Integration tests

We use automated testing to check each code change for issues. This means that the code everyone works on is located in a separate git branch (Torvalds(b)) and has to pass all integration tests before it gets merged into the main branch. A branch can be seen as a copy of the project that one works on in parallel to the original. When the work is done, the changes are copied over to the original version.

But before the changes can be added to the main branch, they have to pass our tests, which are basically dummy-programs that get executed on Travis and AppVeyor and require different parts of the software to work. This assures that we always have a stable version and improves code quality.
These tests are performed on remote servers and the status is visible on GitHub. (GitHub, Inc.(b))

Travis and AppVeyor both use the same technology. They create a virtual machine to simulate a computer and run our software on that machine.

### 5.2.1 Travis

Travis is a german service provider for automating integration tests that can be found under https://travis-ci.org/.

Travis offers its services for free to open source projects. (Travis CI, GmbH(a))
We use it to compile and test our code on Linux. Travis also supports macOS, but since they both use the same compiler we chose to just use Linux.

Travis also generates the PDF's for our documentation and warns us if a citation is missing a bibliography entry.

This automatic generation allows us to control the provided pdf remotely, without the need for building it offline.

The services we used from travis have one big downside - they have no caching or preinstalled configurations. This means when using LaTeX or modern compilers that are still under development and not fully released, they have to be installed first, and this will take its time.

Having the security of knowing when the PDF of the documentation still builds is something we value a lot and have learned to value even more when multiple people work at the same time.

### 5.2.2 AppVeyor

Travis is a canadian service provider for automating integration tests on windows that can be found under https://ci.appveyor.com.

Appveyor also provides its services for free to open source projects. (Appveyor Systems Inc.())
We use it to compile and test our code on Windows with Visual Studio.

We struggled a lot with appveyor, because our Visual Studio configs were based on Visual Studio 2017 RC and they required this version to run.

However, when first used by us, Visual Studio 2017 RC was just released in closed beta as a pre-installation for continous integration. We had to get access to it by requesting for access through the public repository on github.

Once correctly configured however, we never had any problems with appveyor.

### 5.3 CMake

CMake is a tool to control processes of software compilation and testing. (Kiteware())
It allows us to write a fairly simple configuration file which can then be used on multiple platforms. It is a high level configuration that has to be converted into a platform specific one. This conversation is being done automatically by the build system (CMake) and thus does not cost us any time.

This allows us to support almost every operating system, as it was important for us the be platform independent.

The problem with platform dependent solutions is that they are not as accessible to everyone, and we really want to support all major operating systems to make our code and work as accessible as possible.

CMake supports a hirachial setup of its build-tool, that allows you to move parts of the build tool to subfolder, and then chaining the build-scripts together with a root build script.(Kiteware())

## 5.4 Challanges

## 5.5 CLion

CLion is an integrated development environment for C++ developers. (JetBrains s.r.o.(b))
We decided to use it over other available tools because it is the best tool available for macOS and Linux and we felt like we wouldn't get any productivity raise otherwise after trying several other IDEs.

One of the problem with CLion is that it is not up-to-date with all of the latest developments in the C++ programming language. This makes it almost impossible to use it for modern C++ development.

We then even started using plain text editors in edge cases on Linux to not be limited by the editor and compile our code with the new C++ features with compiler that we accessed via the command-line (JetBrains s.r.o.(a))

This was only a problem on Linux and maxOS, because for Windows, the very well known Visual Studio IDE is available, which supports the features we wanted in a release candidate that is publicly available.

# 6 Combining Neuro-Evolution of Augmenting Topologies with Convolutional Neural Networks

## 6.1 Challenges & Solutions

The goal of NEAT is to make topological units modular. These can then be combined in a not predetermined way. So our two questions while combining become:

1. How can we make CNN's modular?

2. How can these units be combined in a meaningful way?

Our first approach was simply taking NEAT and exchanging some of the neurons for Filters. An example network can be seen here:

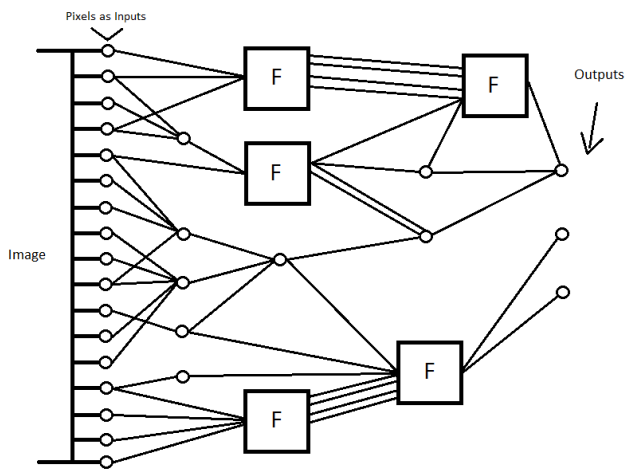

This approach is probably as modular as it gets, however it brings various problems when combining.

1. We ignore one of the main advantages on CNN: Being able to drastically lower the number of inputs through subsampling

2. We don't use Pooling or ReLU layers

3. The significance of a single classic neuron in such a system is questionable

4. The filters in the same layer have to have some way of communicating to form a convolution

5. Adding a new filter in a convolution conflicts with previous learned parameters

We can't address all of these conflicts in a satisfactory way, so we decided to go on to a next approach. We adressed issue 3 by separating the whole network into a convolutional and a fully connected part. This allows us to take issue 1 by adding the concept of a *minimal network*, inspired by NEAT's practice of always starting with combining all inputs with all outputs.

In our case, the minimal network would incorporate some combination of convolution and pooling to reduce the input space. While the exact form of it is debatable, we think a good starting point is LeNet, as it proved itself to be flexible in its application. (Yann LeCun & Haffner(1998))

The overhauled version would start out like this:

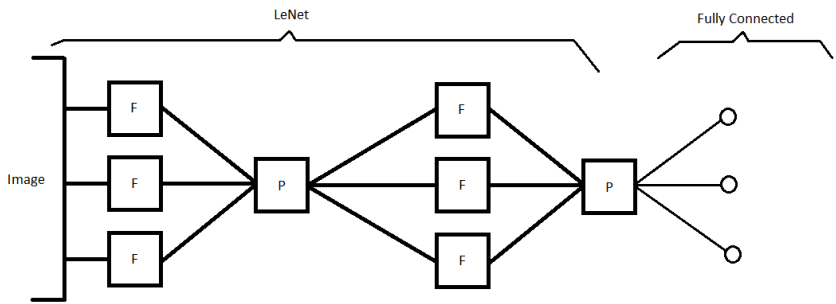

And could evolve into something like this:

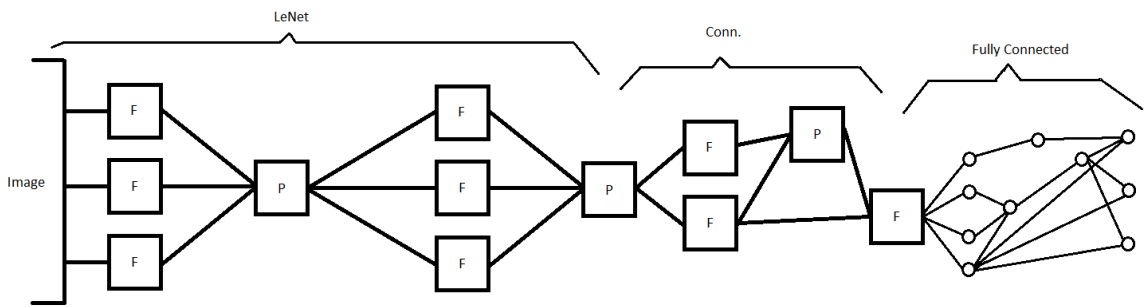

This setup is problematic because filters are supposed to work together to form convolutions.

To process the same input (issue 4), the filters need to have the same size, which we cannot guarantee once we randomly insert new filters or, as per issue 2, pooling layers.

We can only scale the weight matrixes in the filters to the same size by either filling the smaller ones with a bunch of meaningless zeros or pooling the bigger one down, which, beeing a non-lossless compressing algorithm, makes our matrix less accurate.

We come to the conclusion that we have to limit the modularity of the Filters, as doing otherwise brings to many cons. Instead of letting the filters connect to whatever they want, we group them in convolutions. These can alter the dimensionality of all filters in them at once, guaranteeing homogeneity and encapsulation.

With the filters now being synchronized in their convolutions, we have no more problems introducing poolers or ReLUs, as a convolution as a whole doesn't care about the size of it's input matrix.

Our updated pool of available units for stochastic insertion is now:

| **Convolutional** | **Fully connected** |
| --- | --- |
| Convolution | Neuron |
| Pooler | |
| ReLU | |

Our starting topology now looks like this:

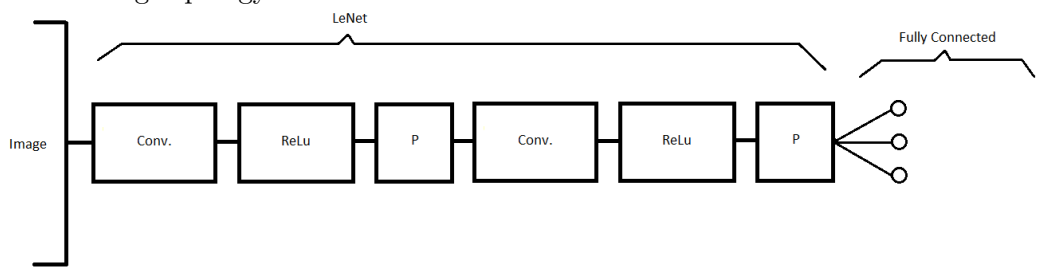

The possible developments consist of a chain of random units right after LeNet.

This raises a new question: *How is the meaning of the fully connected part altered when we add a new unit in the convolutional part?*

After detailed evaluation, we came to the conclusion that all of the parameters in the fully connected part would be fine tuned to a specific expected input. This expectation however ceases to be met once the dimensionality of the convolutions changes, as this shifts a lot of weight parameters towards a new meaning.

This means that we have two choices on how to process the fully connected part in case of a topological change in the convolutional part:

1. Adjust weights for the new meaning

2. Trash the fully connected part and train it anew

Both of these possibilities are not satisfactory. 1 will take a long time, since the already trained fully connected 4structure is basically meaningless now. 2 throws away big, otherwise perfectly usable, parts of the network.

After some research into this problem we found a recent paper from Google, describing how to get rid of the fully connected layer completely by using a global average pooler (Min Lin(2014)).

If we treat the feature map matrix $F$ at the $l$th dimension as a vector $F'_l$, the global average pooler is defined as follows:

$$f(F'_l) = \frac{\sum_{i=1}^n F'_{li}}{n}$$

We then forward the results of every layer to the softmax layer.

Provided the last layer of the convolutional part outputs a tensor with exactly as many dimensions as the number of possible network output, we can exchange the complete fully connected part for this global average pooler while achieving the same results with a drastically improved performance in both evaluation and search space. (Min Lin(2014))

The reason is, in a nutshell, that we stop imagining the output of a filter as detection of a feature.

We now treat it as a rate of confidence: The bigger the numbers, the more confident we are that the feature is present.

This means that the feature detection is no longer performed by the fully connected part, but instead by every single filter in the network together (Min Lin(2014)). Our standard network now looks like this:

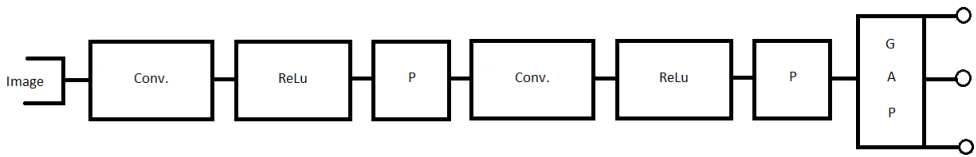

And could develop into something like this:

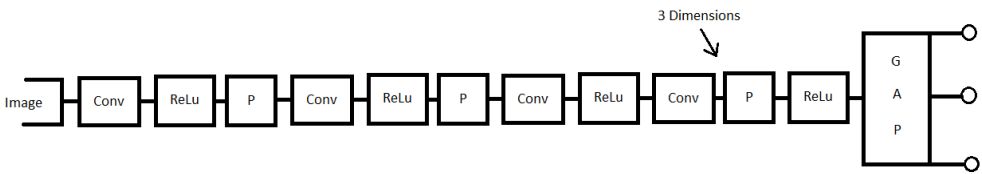

We now seem to have resolved all issues, however when looking at the layers of the example, we see that it has a depth of 8 logical layers (ReLU layers are not counted because they do not result in a feature extraction, as they are merely activation functions). This huge amount is very atypical and has been shown to result is various problems such as very high hardware requirements and lower accuracies(Simonyan & Zisserman(2015)).

The fundamental problem is that the effect of a change in the parameters in a lower layer becomes abysmal compared to a change in the higher ones (Simonyan & Zisserman(2015)) (Hochreiter(1991)). A network of this size is not realistically trainable by us.

A very recent paper now belonging to the Facebook AI Research group deals with these issues.
They introduce the concept of Residual Networks, in short ResNets. (Kaiming He & Sun(2015))
Their goal was to create a convolutional network by combining an arbitrary amount of well defined residual units on which these problems are of no concern. Overly simplified, they address the problem of varying influence by adding a new kind of connecting, called a shortcut.

What it does is simply add matrixes. If they have different dimensionalities, the smaller matrix gets projected on the bigger one by being processed by a one by one matrix with a respective number of filters.

A residual block looks like this:

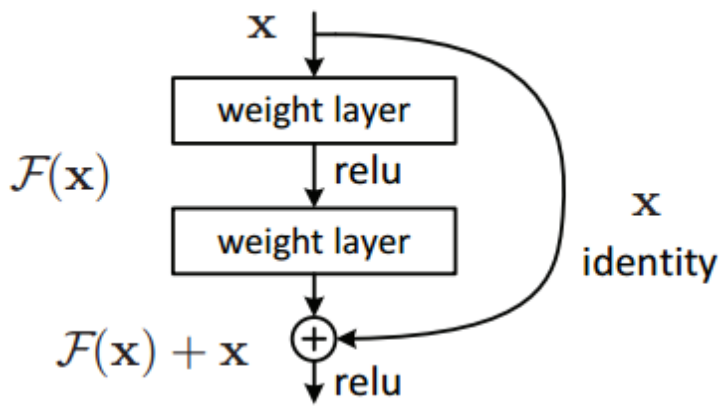

On the left side, a convolutional action takes place (in this case two convolutions with one ReLU activation inbetween). On the right side, the original input of the residual block is added to its output.
This overlay guarantees that the convolutions cannot alter the original state too much, as they now merely highlight features as opposed to extracting them.
The issue of performance is addressed by applying a bottleneck.
This means downsampling the input dimensionality of the residual block by applying one by one con-volutions before performing the convolution and then upscaling it again. This procedure is inspired by Googles Network In Network Inception structure (Kaiming He & Sun(2015)) (Min Lin(2014)).

The overhauled residual block now looks like this:

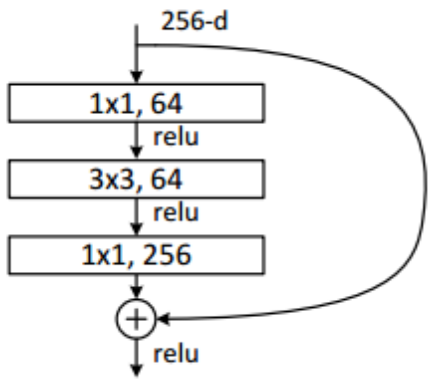

While more convolutions would in theory be possible, only one is used, as the bottleneck dimension poolers introduce new parameters themselves.

This method has been demonstrated to achieve very similar levels of accuracy while reducing a bit chunk of the computational cost (Kaiming He & Sun(2015)).

## 6.2 Definition

Residual Blocks are modular by nature, so they are a perfect fit for our NEAT algorithm.

When analyzing them, we can easily extract following parameters from them:

1. Weights of first dimension pooler

2. Weights of convolution pooler

3. Weights of second dimension pooler

4. Weights of shortcut projection (if needed)

5. Downscaled number of dimensions in each residual block

6. Upscaled number of dimensions in each residual block

7. Number of convolutions in each residual block

8. Total number of residual blocks

Through traditional means we can adjust the parameters 1 to 4.

Numbers 5 to 8 are predefined in ResNet. Their exact values are defined empirically and experimentally. This is of course suboptimal, as we already asserted in chapter two.

We think NEAT can optimize these by encoding them as genes in the genome.

However, because of the nature of our smallest building blocks, it doesn't make sense to store these genomes in a per-connection basis.

All parameters can be described as state of a residual block. For the last one, we just abstract it as a link to the next block. If the algorithm decides to add a new residual block, it can be inserted in a random existing link.

For the parameter tuning, we treat numbers 1 to 4 as a big vector of weights inside the genome of the residual block and apply the same chances and rules of change to them as in standard NEAT , which are:

- Chance of selecting this genome to change weights: 80%

- Chance for each weight to be uniformly perturbed: 90%

- Chance for each weight to be set to a random new value: 10%

(Stanley(2002))

Parameters 5 to 7 are more critical, as they greatly effect the computational cost.

We limited changes to be only +1 in each, to go with NEAT's thought of starting with a small topology and only going up if necessary. The chances are thus taken straight from how NEAT treats extra neurons: Each genome has a 3% chance of mutating one of the mentioned parameters at birth.

Lastly, parameter 8 is the one directly in control of the network's depth. As ResNet proved, deep networks have great advantages to shallow ones (Kaiming He & Sun(2015)). So we made this value very prone to grow by one. The chance is analog to NEAT's chance of adding a new connection, which is 30% at birth.

Additionally, we changed the compatibility functions $c_3$ parameter to 0.06 to account for the higher number of potential weight differences per genome.

## 6.3   Implementation

While programming according to our algorithm, we continuously tested our code.
We unfortunately found out during one of those tests, that our implementation of matrix multiplication while applying a filter is not nearly fast enough to process high quality scans of mammographies.

A simple test with 32 by 32 images confirmed our fear: Deep networks with a dimensionality higher than 10 are not realistically computable in a given time. By contrast, the deepest ResNet uses more than 1000 dimensioniona in its lowest layers. Given that genetic algorithms go even further by not training one network, but 100 at a time, and considering our limited time, we had to halt further research.

# 7 Further enhancements

## 7.1 Optimisation

The single biggest challenge we faced was performance.

We had estimated that for training a full set of 800 pictures at a mere 400 to 400 pixels, we would need months for just training the network once. This held us back from efficiently mesuring our algorithms correctness.

> *"Currently, large-scale CNN experiments require specialized hardware, such as NVidia GPUs, and specialized APIs, such as NVidia's CuDNN library, to achieve adequate training performance."* (Abuzaid(2015))

Firaz Abuzaid also mentions that "at runtime, the convolution operations are computationally expensive and take up about 67% of the time; other estimates put this figure around 95%".

We were (unfortunately) able to confirm these numbers as realistic - one line of code (the multiplication of the matrices values) took up 93% of the execution time when testing our code, the loop for executing these multiplications took another 6% of the execution time.

Here are some improvements that could be done to optimise the performance of convolutional neural networks:

- Using GPUs to accelerate matrix-multiplications(Hochberg(2012))

  Using the power of GPUs for complex and computation heavy calculations has been very important the last years in the industry. GPU toolkits seem to consistently perform the same tasks five to ten times faster than their CPU counterparts.(Abuzaid(2015))

- Using the CcT method to optimize CPU usage

  The CcT method has proven to be up to 4 times faster than one of the often used CPU toolkits for machine learning; Caffe. Utilizing this method would allow us to improve the performance of CNNs by a big margin without having to use expensive GPUs.(Abuzaid(2015))

There are other approaches of optimizing CNNs to be more efficient, such as Low Rank Expansions(Max Jaderberg(2014)), the approach of Optimizing a FPGA-based Accelerator Design for Deep Convolutional Neural Networks(Chen Zhang(2015)) and Convexified Convolutional Neural Networks(Yuchen Zhang(2016)).

All these approaches share the same limitations for us - it is unclear whether they are even compatible with our NEAT based evolutionary algorithm, and if they are, the changes to the inner workings of our algorithm would be drastic, so that benchmarking would be hard. Due to the recency of these developments, it is hard to fully estimate their impact onto our model and performance.

## 7.2 Safety concerns

We learned in the presentation of Dr. Krause at the SGAICO Annual Meeting and Workshop - Deep Learning and Beyond in Luzern of the concept of safety constraints.
He offered insight into his current studies about how to train system that have influence over real life and can cause harm. Examples where:

- A quadcopter learning to fly around a stationary object. It could potentially fly in a manner resulting in a crash, damaging itself or propriety, causing financial damage.

- A system learning how to apply a new experimental treatment to patients. This can end in life threatening circumstances.

A big point to consider here is the bayesian concept of false positives vs false negatives. In other words: "What is more critical, telling a patient he is sick when he is not (false positive) or telling him he is fine when he is actually pretty ill (false negative)?"
Of course, the answer to that depends on multiple factors such as treatment cost and lethality of the condition. Dr. Krause proposes mechanisms that do not allow damaging decisions once you have settled on a definition of what "damaging" means in context of the training. We think this is very relevant to the field of medical diagnostics and so a good improvement to consider in the future.

## 7.3 HyperNEAT

HypernNEAT is a further adjusted version, often also called an extension, of the original NEAT algorithm.(Jessica Lowell & Grabkovsky(2011))

HyperNEATs major problem is that it has performance hits compared to the original version of NEAT. With that, the already performance flawed system of combining NEAT with CNNs would be too slow:

> *"Finally, one major problem with HyperNEAT is that it is very slow, even on a multi-core processor."*(Jessica Lowell & Grabkovsky(2011))

# 8 Our Work

We have dedicated our project work to the subject Image recognition by artificial intelligence.

Image recognition by artificial intelligence has and interests us very much, because you can create something that does not exists yet. Our project is mainly concerned with computer science (artificial intelligence) and medicine. An important aspect for our work was to create something that can be needed in the future and what can be of benefit to other people. Because we are software engineers, it is also a good opportunity to train us in our area.

## 8.1 Collaborators

### 8.1.1 Project Group

Mr. Ferner has thought of realizing this project quite a while ago. Mr. Stucki, Mr. Fischler and Ms. Zarubica have found mutual interest in this topic and wanted to form a group meant to help Mr. Ferner to implement the idea.

Our motivation is to create something new together that can help people with their lives.

### 8.1.2 Acknowledgements

Mr. Benno Piller was the administrative supervisor of the project and has helpfully advised us whenever we had administrative questions or were in need of an external opinion.

Ms. Polina Terzieva, Bachelor of English philology, proofread multiple sections and provided sporadic support in linguistic and stylistic questions.

### 8.1.3 Medical Support

We have contacted two medical specialists who are willing to look at our project and help us with it by providing us with data.

They are:

PD Dr. med. univ. Christoph Tausch

General surgeon with focus on clarification and treatment of breasts

Brustzentrum Zürich

Dr. med. Serafino Forte

Deputy ead doctor radiology

Kantonsspital Baden

Both have submitted a request for a studyprotocol, which can be found in the attachments.

Dr. Tausch also wanted to know about the type of data needed (which age, gender, cancer type, etc.)

## 8.2 Our goals

While working on our project we had to take our goals into account. So we found a lot oft them. Here are our main goals.

- Read mammography correctly

- Probability indication to 95%

- Visual display

- Present the knowledge to the layman

- Reach platform independency

The ability to read mammography is very important because besides the project consisting of image recognition, it needs mammography too. Certainty is also really important for the project, and it puts a "stamp" on it because one has to be very sure before putting a cancer diagnose.

A visual display is also a goal the project has, as it makes the use of the software easier and more appealing to the eye. A presentation for the layman could be a good supplement, providing them with the information they need, just to have a general clue.

As a conclusion, our project should be softwate-independent with the purpose that one could be able to use it on any device.

## 8.3  Initial position

- The project took off with a semi-finished NEAT-library
- There is a library named "Hippocrates" consisting of approximately a half a year of work

## 8.4  Opening questions

- Is it possible to combine NEAT and CNN?

- Is it possible to carefully explain the complexity of subjects and if possible, simplify them?

- Is it possible to cooperate and work with the hospital?

- Is it possible to evaluate mammographys at home with the software we possess?

- Is our software capable to emulate a human evolution?

- Is our software usable as an assistance system in the future?

## 8.5  Working programms and tools

We have used the following platforms and programs for the work Visual Studio; LaTex; and Github.

### 8.5.1  Visual Studio

Visual Studio is a programming tool developed by Microsoft for Windows we have programmed all of our software in. We have worked with it quite a lot as well. Since Visual Studio is very modern, it didn't cause us problems and without a doubt helped us with the project changes that we made.
Changes applied to the software structure were not too difficult.

### 8.5.2 LaTex

The tool we used for our documentation and design is LaTex, the advantages of which are that LaTex takes complete care of the presentation of the documentation, including the citations. This guarantees that we waste as little time as possible with visually designing the documentation.

Some rather special formulas can also be used wich the program formats it by itself so we don't have to worry about it.

### 8.5.3 Github

Github is an online platform we used to store the code we have written in. In Github people can work on projects together and view other people's work. It can also work in a way that people give other people tasks or demand a review of a task. Github is and has been an enormous help to our project, beause it has given us an overview of the things we have programmed. On top of that, Github gives us reasssurement because every work has been checked and reviewed by some other member of the site.

In Github you can also make branches. A branch is a copy of a project. On one hand, it offers additional security for the programmer because if any changes are made in the original project, it can be possible that this would no longer works. On the other hand, if one is working on the branch, nothing can be destroyed.

Changes and edits in the branch can of course be made in the original project, but only if it's in order. Since our project was public on Github, people who had no idea about it, but were interested in it, could take a look at it.

Specialists in our area of expertise could look at our program and give us tips, for example on how to improve it. Github has many advantages and offers the user many possibilities.

## 8.6 Procedure

### 8.6.1 The beginning

Our starting point was picking a topic for the project.

We haven't been dealing with this for quite a long time because we already had a concrete idea of what the project was going to be about.

Jan Hohenheim (project member) had an inspiring idea with which he truly inspired us.

Roughly one year before the project he had the idea to make possible the recognizing of different images using artificial intelligence.

In the course of this year, he collected information and experiences, and came up with a concept to make image recognition possible in the field of medicine, more precisely breast cancer (mammography).

When we started out the project, we had a very interesting topic and also someone who had previously dealt with it.

### 8.6.2 The planning

At the beginning, a rough plan of how the project is going to develop was made.

We have looked at the aspects which are very time-consuming and important. After this, we proceeded by making a weekly planning.

We knew what we had to do and how to do it, of course thanks to the planning but since that wouldn't be enough, we had to group each week and discuss tasks carefully.

We also discussed how we would like to cooperate with the doctors.

### 8.6.3 The realisation

Since the planning has been completed, it was time for the realization.

As already mentioned in the planning, we worked in a weekly schedule.

This means that we divided all the big tasks into smaller ones, each one going on for about a week.

Having a really good overview of our project, we knew exactly where we stand.

Also we had a meeting every week and discussed our concerns about the project.

The biggest problem with the realization was that we lacked the computing power with which to run our project and visualize it.

We had a problem to present our project correctly and to make it clear to the people what we have achieved completing it.

Apart from this, the realization ran very well.

Our job was to mainly work on NEAT (NeuroEvolution of Augmenting Topologies) and CNN (Convolutional Neural Network).

When both parts were completed, it was our goal to connect them together.

In addition, we cooperated with a few doctors that helped us by providing us with data we needed.

We searched for them on the internet and in books, at least the ones that were specialists in the field of oncology (tumor diseases).

We prepared a study protocol to present our project to the doctorsd.

After checking out our study protocol, Dr. Serafino Forte from the Hospital of Baden invited us to introduce him to our project.

He gave us some helpful tips and we discussed the next steps.

### 8.6.4   The result

The result led to a dichotomy.

The two largest parts, CNN (Convolutional Neural Network) and NEAT (NeuroEvolution of Augmenting Topologies), have been successfully connected together and they worked successfully. However, our computing power is much too low to test our software on.

One of our main concerns is to explain people who haven't encountered our program and this type of work that it would still run successfully, despite us not being able to show it to them or visualize it.

With the performance of a really good laptop, the learning of the network would still take about a half a year, which is unfortunately time we don't have.

### 8.6.5   Our conclusion

Our software worked successfully and we are very satisfied with it. Unfortunately, we have used an outdated technology of mammography, which is nowadays almost to no longer needed. So it turns out that our software can serve in only a few facilities as an assistance system. However, this doesn't necessarily mean it's negative, quite the contrary - an advantage. Inserting the new technology is definitely possible, and therefore our software would also support all technologies.

In conclusion, our software can be used as an assistance system, but we want to optimize the data to make it easier for the doctors to work with it and eventually be able to help people thanks to the provided information.

The project was a very good opportunity to let our ideas run free and to further train our knowledge. We are very optimistic for our software to be used and in future optimized.

## 8.7 Progress

Progress made until the end of October

Implement the load of networks (ZAR)

Complete the visualizer (MFI)

Complete the documentation part "How does a neural network work?" (JST)

Progress made until the middle of November

Complete the study journal (ZAR)

Complete CNN (JNF, JST)

Complete NEAT (ZAR, MFI)

Progress made until the end of November

Complete the documentation part "Automatic testing of software" (JST)

Complete the documentation part "What is NEAT?" (MFI)

Progress made from the middle of November until the middle of December

Connect CNN with NEAT (All)

Testing of different variations (All)

Execute benchmarks (All)

Possible tests in and cooperation with other people (All)

Progress made from mid to end of December

Evaluation of tests, writing of conclusions (All)

Checkpoints

16.11.2016: SGAICO Meeting- Deep Learning and Beyond

23.11.2016: Delivery of the fine concept (Table of contents)

18.01.2017: Delivery of the project and presentation

25.01.2017: Project exhibition

## 8.8  Contact with doctors

Not much time has passed after the planning until we already began looking for doctors who could cooperate with us on the project.

For this purpose we sought doctors who were active exclusively in the field of oncology.

The term oncology is reffered to the branch of medicine that deals with the diagnosis, treatment and prevention of different types of cancer (tumor disease).

With our project being almost entirely about breast mammography, we needed doctors which are specialized in the field of oncology, more specific - breast oncology. According to this categorization we then proceeded.

After some research on the internet and also in books, we finally cut down 26 doctors who could be considered qualified to for the project.

We have sent all of the doctors a brief description of our project. We explained to them the core and the content of the project. Some of them answered the e-mails we sent.

Most of the doctors have too much research to do, thus they just don't have as much time.

We arranged a few appointments, which were unfortunately not in the time span of our project.

Two doctors sent us a study protocol request, which means they required a protocol of our project that is extremely detailed.

We wrote the following protocol and sendt it to the doctors.

This we sent the doctors.

Dr. Serafino Forte of the hospital of Baden has invited us to the hospital after examining the study protocol.

We talked to him about all the details in medical and organizational terms. At first he looked at our mammography record. Immediately he realized that our mammographys are obsolete, because this technology is no longer needed and probably outdated.

Our data was in the old format of the film-film system. This was good for our project on the one hand but on the other hand it didn't speak so good.

It is definitely an advantage, because we have the possibility to have different mammographys evaluated.

However, it has a disadvantage too.

It is very difficult to get to the new mammographys because we need a lot of mammographys to train our software, we have to submit an ethics application.

This is an even more detailed application than the study protocol, which is checking if our project is as trustworthy as possible.

The application is examined by the Swiss Ethics Committees for Human Research.

In this application the financial resources are examined.

The financial resources are checked because the application is not free of charge.

To submit an application, you have to pay a basic flat rate of 800.00 CHF.

A check should be made for how well can the people be put together. This means that four individuals working as a team could possibly carry out such a project because at the end of it it's just a product, resulting from months of work.

Apart from the organizational, we have also looked at the medical aspects as mentioned earlier in the chapter.

For Dr. Serafino Forte, it was quite important that our software could not only tell you whether a person is diagnosed with cancer, but also what type of cancer does he have.

# References

Abuzaid(2015). Abuzaid, F. (2015). Optimizing cpu performance for convolutional neural networks.
URL http://cs231n.stanford.edu/reports/fabuzaid_final_report.pdf

Alex Krizhevsky & Hinton(2012). Alex Krizhevsky, I. S., & Hinton, G. E. (2012). Imagenet classification with deep convolutional neural networks.
URL https://papers.nips.cc/paper/4824-imagenet-classification-with-deep-convolutional-neural-networks.pdf

Anderson(1995). Anderson, J. (1995). *An Introduction to Neural Networks*. MIT Press.

Antwerpes(2015). Antwerpes, D. F. (2015). Endokrinologie.
URL http://flexikon.doccheck.com/de/Endokrinologie#

Appveyor Systems Inc.(). Appveyor Systems Inc. (????). Appveyor plans and pricing.
URL https://www.appveyor.com/pricing/

Bäck(1996). Bäck, T. (1996). Evolutionary algorithms in theory and practice. *Oxford Univ. Press*.

Beams(2014). Beams, C. (2014). How to write a git commit message.
URL http://chris.beams.io/posts/git-commit/

Brownlee(2016). Brownlee, J. (2016). How to implement the backpropagation algorithm from scratch in python.
URL http://machinelearningmastery.com/implement-backpropagation-algorithm-scratch-python/

Buckland(). Buckland, M. (????). Genetic algorithms in plain english.
URL http://www.ai-junkie.com/ga/intro/gat1.html

Chen Zhang(2015). Chen Zhang, G. S. Y. G. B. X. J. C., Peng Li (2015). Optimizing fpga-based accelerator design for deep convolutional neural networks.
URL https://pdfs.semanticscholar.org/2ffc/74bec88d8762a613256589891ff323123e99.pdf

Cowan(2014). Cowan, M. K. (2014). neural.
URL https://github.com/battlesnake/neural/tree/7dd93c49527ce3ff3621d09c7fa6369411901f76

GitHub, Inc.(a). GitHub, Inc. (????a). About pull request reviews.
URL https://help.github.com/articles/about-pull-request-reviews/

GitHub, Inc.(b). GitHub, Inc. (????b). Continuous integration.
URL https://github.com/integrations/feature/continuous-integration

Graham(2014). Graham, B. (2014). Computer vision and pattern recognition (cs.cv).
URL https://arxiv.org/abs/1412.6071

Green(2009). Green, C. D. (2009). Speciation in canonical neat.
URL http://sharpneat.sourceforge.net/research/speciation-canonical-neat.html

Haşim Sak(2014). Haşim Sak, F. B., Andrew Senior (2014). Long short-term memory based recurrent neural network architectures for large vocabulary speech recognition.
URL https://arxiv.org/abs/1402.1128

Hess(2011). Hess, B. (2011). *Publicus 2012*. Schwabe AG.

Hinton(2007). Hinton, G. E. (2007). Learning multiple layers of representation.
URL http://www.cs.toronto.edu/~fritz/absps/tics.pdf

Hochberg(2012). Hochberg, R. (2012). Matrix multiplication with cuda — a basic introduction to the cuda programming model.
URL https://www.shodor.org/media/content/petascale/materials/UPModules/matrixMultiplication/moduleDocument.pdf

Hochreiter(1991). Hochreiter, S. (1991). Untersuchungen zu dynamischen neuronalen netzen.

Jason Yosinski & Lipson(2015). Jason Yosinski, A. N. T. F., Jeff Clune, & Lipson, H. (2015). Understanding neural networks through deep visualization.
URL http://yosinski.com/deepvis

Jessica Lowell & Grabkovsky(2011). Jessica Lowell, K. B., & Grabkovsky, S. (2011). Comparison of neat and hyperneat on a strategic decision-making problem.
URL http://web.mit.edu/jessiehl/Public/aaai11/fullpaper.pdf

JetBrains s.r.o.(a). JetBrains s.r.o. (????a). C++ support.
URL https://www.jetbrains.com/help/clion/2016.3/cpp_support.html

JetBrains s.r.o.(b). JetBrains s.r.o. (????b). Clion.
URL https://www.jetbrains.com/clion/

Kaiming He & Sun(2015). Kaiming He, S. R., Xiangyu Zhang, & Sun, J. (2015). Deep residual learning for image recognition.
URL https://arxiv.org/pdf/1512.03385v1.pdf

Kaiming He & Sun(2016). Kaiming He, S. R., Xiangyu Zhang, & Sun, J. (2016). Identity mappings in deep residual networks.
URL https://arxiv.org/pdf/1603.05027v3.pdf

Karen Simonyan(2014). Karen Simonyan, A. Z., Andrea Vedaldi (2014). Deep inside convolutional networks: Visualising image classification models and saliency maps.
URL https://arxiv.org/pdf/1312.6034v2.pdf

Karpathy(2016). Karpathy, A. (2016). Cs231n convolutional neural networks for visual recognition.
URL http://cs231n.github.io/convolutional-networks/

Kiteware(). Kiteware (????). Cmake.
URL https://cmake.org

Masakazu Matsugu(2003). Masakazu Matsugu, Y. M. Y. K., Katsuhiko Mori (2003). Subject independent facial expression recognition with robust face detection using a convolutional neural network.
URL http://www.iro.umontreal.ca/~pift6080/H09/documents/papers/sparse/matsugo_etal_face_expression_conv_nnet.pdf

Max Jaderberg(2014). Max Jaderberg, A. Z., Andrea Vedaldi (2014). Speeding up convolutional neural networks with low rank expansions.
URL https://www.robots.ox.ac.uk/~vedaldi/assets/pubs/jaderberg14speeding.pdf

Microsoft(2016). Microsoft (2016). Interop marshaling.
    URL `https://msdn.microsoft.com/en-us/library/eaw10et3(v=vs.110).aspx`

Min Lin(2014). Min Lin, S. Y., Qiang Chen (2014). Network in network.
    URL `https://arxiv.org/pdf/1312.4400v3.pdf`

Nguyen & Widrow(1990). Nguyen, D., & Widrow, B. (1990). Improving the learning speed of 2-layer
    neural networks by choosing initial values of the adaptive weights.
    URL `http://www-isl.stanford.edu/~widrow/papers/c1990improvingthe.pdf`

Nielsen(2016). Nielsen, M. (2016). Neural networks and deep learning.
    URL `http://neuralnetworksanddeeplearning.com/chap6.html`

nzhagen(2016). nzhagen (2016). bibulous.
    URL
    `https://github.com/nzhagen/bibulous/tree/5bbbe39ed313c1cfd531a40437d8a56f35694c68`

Philipp Krähenbühl(2016). Philipp Krähenbühl, J. D. T. D., Carl Doersch (2016). Data-dependent
    initializations of convolutional neural networks.
    URL
    `https://ai2-s2-pdfs.s3.amazonaws.com/8e56/448da09ceacea946c9d6fd393ad3e57e12cb.pdf`

Polamuri(2014a). Polamuri, S. (2014a). Supervised and unsupervised learning.
    URL `http://dataaspirant.com/2014/09/19/supervised-and-unsupervised-learning/`

Polamuri(2014b). Polamuri, S. (2014b). Supervised and unsupervised learning.
    URL `http://dataaspirant.com/2014/09/19/supervised-and-unsupervised-learning/`

Preston-Werner(a). Preston-Werner, T. (????a). Github.
    URL `https://github.com`

Preston-Werner(b). Preston-Werner, T. (????b). Github pricing.
    URL `https://github.com/pricing`

Rojas(1996). Rojas, R. (1996). Neural networks.
    URL `https://page.mi.fu-berlin.de/rojas/neural/neuron.pdf`

Sathyanarayana(2014). Sathyanarayana, S. (2014). A gentle introduction to backpropagation.
    URL `http://numericinsight.com/uploads/A_Gentle_Introduction_to_Backpropagation.pdf`

Simonyan & Zisserman(2015). Simonyan, K., & Zisserman, A. (2015). Very deep convolutional networks
    for large-scale image recognition.
    URL `https://arxiv.org/pdf/1409.1556v6.pdf`

Stanley(2002). Stanley, K. (2002). Evolving neural networks through augmenting topologies.
    URL `http://nn.cs.utexas.edu/downloads/papers/stanley.ec02.pdf`

Stanley(2010). Stanley, K. (2010). Neat c++.
    URL `http://nn.cs.utexas.edu/?neat-c`

Strassner(2015). Strassner, T. (2015). Xml vs json.
    URL `http://www.cs.tufts.edu/comp/150IDS/final_papers/tstras01.1/FinalReport/FinalReport.html`

Stroustrup(2013). Stroustrup, B. (2013). *The C++ Programming Language (4th Edition)*. Addison-Wesley.

The NetBSD Foundation(). The NetBSD Foundation (????). Netbsd ftp server.
URL `http://ftp.fi.netbsd.org/pub/graphics/packages/mpeg/havefun.stanford.edu/cv/`

Torvalds(a). Torvalds, L. (????a). git.
URL `https://git-scm.com`

Torvalds(b). Torvalds, L. (????b). git branch.
URL `https://git-scm.com/docs/git-branch`

Torvalds(c). Torvalds, L. e. a. (????c). Getting a git repository.
URL `https://git-scm.com/book/en/v2/Git-Basics-Getting-a-Git-Repository`

Travis CI, GmbH(a). Travis CI, GmbH (????a). Travis ci plans.
URL `https://travis-ci.com/plans`

Travis CI, GmbH(b). Travis CI, GmbH (????b). Travis ci plans.
URL `https://travis-ci.com/plans`

University of South Florida(). University of South Florida (????). Digital database for screening mammography.
URL `http://marathon.csee.usf.edu/Mammography/Database.html`

van den Branden Lambrecht(2001). van den Branden Lambrecht, C. J. (2001). *Vision Models and Applications to Image and Video Processing*. Springer Science & Business Media.

Yann LeCun & Haffner(1998). Yann LeCun, Y. B., Leon Bottou, & Haffner, P. (1998). Gradientbased learning applied to document recognition.
URL `http://vision.stanford.edu/cs598_spring07/papers/Lecun98.pdf`

Yuchen Zhang(2016). Yuchen Zhang, M. J. W., Percy Liang (2016). Convexified convolutional neural networks.
URL `https://pdfs.semanticscholar.org/165f/b135ffbaa1ab63bc4e59dc2bbc8f5ea7bfdc.pdf`